\documentclass{article}

\PassOptionsToPackage{numbers, compress}{natbib}
\usepackage[preprint]{neurips_2026}

\usepackage[utf8]{inputenc}
\usepackage[T1]{fontenc}
\usepackage{hyperref}
\usepackage{url}
\usepackage{graphicx}
\usepackage{booktabs}
\usepackage{amsfonts}
\usepackage{amsmath}
\usepackage{amssymb}
\usepackage{nicefrac}
\usepackage{microtype}
\usepackage[table]{xcolor}
\usepackage{multirow}
\usepackage{makecell}
\usepackage{wrapfig}
\usepackage{xcolor}

\definecolor{cogmapteal}{RGB}{43, 173, 160}

\title{SpaceMind++: Toward Allocentric Cognitive Maps for Spatially Grounded Video MLLMs}

\author{%
  \normalfont 
  \textbf{Bo Gu$^{1\dagger}$, Zhikang Zhang$^{2\dagger}$, Zizhuang Wei$^{2}$, Zhenyuan Chen$^{2}$} \\
  \textbf{Lingyun Li$^{2*}$, Zhuoyi Song$^{3,1*}$} \\[0.5em] 
  $^{1}$Fudan University \quad
  $^{2}$Huawei \quad
  $^{3}$Shenzhen Loop Area Institute \\
  \small $\dagger$ Equal contribution \quad $*$ Corresponding author
}

\begin{document}

\maketitle

\begin{abstract}
Recent multimodal large language models (MLLMs) have made remarkable progress in visual understanding and language-based reasoning, yet they lack a persistent world-centered representation for spatially consistent reasoning in 3D environments. 
Inspired by the mammalian dual-stream system, where semantic and spatial cues are processed separately and integrated into an allocentric cognitive map, we propose \textit{SpaceMind++}, a video MLLM architecture that explicitly builds a voxelized cognitive map from RGB videos. This map reorganizes fragmented egocentric observations into a shared 3D metric representation, enabling the model to preserve object permanence and spatial topology across changing viewpoints. To make this allocentric representation usable by a pretrained video MLLM without disrupting its native visual-token interface, we introduce Coordinate-Guided Deep Iterative Fusion, a new mechanism that relays map-level spatial knowledge back into the original 2D visual features. This fusion is explicitly guided by coordinate embeddings and 3D Rotary Positional Encoding, which ground semantic interactions in metric 3D space, resembling the entorhinal binding of sensory features to metric space. Extensive experiments show that SpaceMind++ achieves new \textbf{state-of-the-art} performance on VSI-Bench. Furthermore, it demonstrates superior \textbf{out-of-distribution generalization} on SPBench, SITE-Bench, and SPAR-Bench, underscoring its robustness in unseen 3D environments.

\end{abstract}

\section{Introduction}
Spatial intelligence requires models to reason about the 3D structure of environments beyond isolated visual observations. Despite recent advances that have endowed Multimodal Large Language Models (MLLMs) with strong visual understanding and reasoning capabilities\cite{bai_qwen25-vl_2025,li_llava-onevision_2024,lin_video-llava_2024,zhu_internvl3_2025}, a fundamental gap persists between language-level reasoning and 3D spatial organization\cite{wang_site_2025,zhang_flatland_2025}. This disconnect limits the ability of MLLMs to ground their predictions in physical space, leading to unstable spatial reasoning, such as inconsistent distance estimation and object-level hallucinations\cite{yang_thinking_2024}.
This limitation stems from a lack of spatial constancy: visual observations are egocentric, partial, and transient, with viewpoint-dependent distances, occlusions, and fragmented frame-level evidence of the underlying 3D scene.

Recent efforts to build 3D-aware MLLMs attempt to address this issue by incorporating feed-forward geometry encoders alongside general-purpose visual encoders. These methods commonly fuse 2D visual features and 3D geometric cues through MLPs\cite{wu_spatial-mllm_2025}, cross-attention\cite{fan_vlm-3r_2026}, or camera-guided integration\cite{zhao_spacemind_2025}. However, they largely treat geometry as an auxiliary modality aligned with egocentric visual tokens, enhancing spatial cues only locally without forming an explicit and persistent 3D structure. Rather, reliable spatial intelligence requires more than extracting geometric cues from individual observations; it requires a stable world-centered representation that preserves object permanence, metric consistency, and global scene layout across changing viewpoints.

Human spatial cognition offers a blueprint for spatially grounded reasoning. 
Neuroscience studies highlight two key mechanisms for spatial awareness: the dual visual stream~\cite{goodale_separate_1992,ungerleider_object_1982} and the cognitive map~\cite{hafting_microstructure_2005,okeefe_hippocampus_1978,tolman_cognitive_1948}. The former separates visual processing into ventral “what” and dorsal “where” pathways, while the latter provides an allocentric representation for reasoning about spatial relationships beyond the current field of view. Although cognitive maps have inspired computational models such as the Tolman-Eichenbaum Machine~\cite{whittington_tolman-eichenbaum_2020,whittington_relating_2021,whittington_cognitive_2022} and visual-language navigation systems~\cite{gupta_cognitive_2017,ruan2025reactive}, their use as an architectural principle for large-scale MLLMs remains under-explored.

Inspired by principles of human spatial cognition, we present \textbf{SpaceMind++}, a video MLLM that introduces an allocentric cognitive map as a first-class architectural component for spatial reasoning. In contrast to prior 3D-aware MLLMs that treat 3D geometry as an auxiliary feature stream, SpaceMind++ projects frame-level 2D semantic features into a voxelized map organized in 3D space. This transformation converts fragmented egocentric observations into a structured world-centered representation, allowing the model to maintain spatial topology across viewpoints. 

To make this map useful for language reasoning, we further introduce \textbf{Coordinate-Guided Deep Iterative Fusion (CDIF)}, which relays spatial knowledge from the map back to the original visual tokens through iterative map self-attention and map-to-visual cross-attention. 
The LLM therefore receives geometry-enhanced visual tokens while preserving the pretrained visual-language interface. A key requirement for such fusion is that semantic interactions must be governed by metric 3D relationships rather than by arbitrary sequence order.  
Inspired by hippocampal--entorhinal spatial coding, which supports metric and allocentric representations of space~\cite{hafting_microstructure_2005,rolls_spatial_1999,rolls_spatial_1997},
we make attention coordinate-aware by injecting explicit 3D coordinate embeddings and applying 3D Continuous Rotary Position Embeddings.
This design binds visual semantics to spatial locations during both attention steps, enabling SpaceMind++ to reason over object relations within a 3D metric structure. Experiments on VSIBench and other benchmarks demonstrate that SpaceMind++ achieves state-of-the-art performance in a wide range of visual-based spatial understanding and reasoning tasks.

In summary, our main contributions are:
\begin{itemize}
    \item We introduce a brain-inspired design perspective for spatially grounded video MLLMs, drawing motivation from the mammalian dual visual stream and cognitive map to separate semantic recognition, spatial localization, and allocentric scene organization.

    \item We propose SpaceMind++, a novel video MLLM architecture that builds a voxelized allocentric cognitive map from RGB videos, effectively preserving spatial topology and object permanence across diverse viewpoints.
    
    \item We introduce Coordinate-Guided Deep Iterative Fusion (CDIF), a coordinate-aware mechanism that performs iterative map reasoning and map-to-visual reading with explicit 3D coordinate embeddings and 3D RoPE, while keeping the pretrained MLLM interface intact.
    
    \item We curate SpaceMind-900K for spatial instruction tuning and validate SpaceMind++ across multiple benchmarks, achieving new state-of-the-art performance on VSI-Bench and strong generalization on SPBench, SITE-Bench, and SPAR-Bench.

\end{itemize}

\section{Related Work}

\subsection{Spatial Reasoning in Multimodal Large Language Models}
Multimodal large language models (MLLMs) have achieved strong progress in visual question answering, image captioning, and video understanding\cite{alayrac_flamingo_2022,bai_qwen-vl_2023,chen_internvl_2024,li_blip-2_2023,lin_video-llava_2024,radford_learning_2021,zhu_internvl3_2025}, yet they remain unreliable on spatially grounded tasks such as metric estimation, spatial relationship judgment, and navigation-oriented reasoning\cite{stogiannidis_mind_2025,yang_thinking_2024,yu_sibench_2025}. Aiming to solve this problem, existing spatial MLLMs can be broadly grouped into two lines. One line introduces explicit 3D inputs, including point clouds\cite{fu_scene-llm_2025,hong_3d-llm_2023,mao_spatiallm_2025} and depth maps\cite{cai_spatialbot_2025,cheng_spatialrgpt_2024,zhou_roborefer_2025}, and aligns these structured scene representations with language models. Although effective when high-quality geometry is available, these methods often rely on reconstruction pipelines, limiting their scalability to unconstrained RGB videos. Another line improves RGB-based spatial reasoning by augmenting VLMs with 3D geometry encoders or internal spatial representations, as explored in Spatial-MLLM\cite{wu_spatial-mllm_2025}, VLM-3R\cite{fan_vlm-3r_2026} and SpaceMind\cite{zhao_spacemind_2025}. These approaches demonstrate the value of geometric priors, but typically integrate spatial information through token-level fusion. Such designs enhance local spatial cues but do not explicitly reorganize observations into a persistent world-centered structure. Our work instead treats 3D geometry as the organizing substrate for constructing an allocentric cognitive map, enabling more stable spatial reasoning across changing viewpoints.

\subsection{Feed-forward Visual Geometry for 3D Scene Understanding}
Classical 3D reconstruction pipelines, such as Structure-from-Motion~\cite{schonberger_structure--motion_2016} and Multi-View Stereo~\cite{schonberger_pixelwise_2016,wang_patchmatchnet_2021,yao_mvsnet_2018}, recover camera poses and dense geometry from images. However, their reliance on multi-stage optimization and sufficient view overlap makes them difficult to integrate into end-to-end multimodal reasoning systems. 
Recent feed-forward visual geometry models, including DUSt3R\cite{wang_dust3r_2024}, MASt3R\cite{leroy_mast3r_2024}, CUT3R\cite{wang_cut3r_2025}, and VGGT\cite{Wang2025VGGTVG}, alleviate this limitation by directly predicting dense correspondences, point maps, depth, and camera parameters from RGB images or short sequences. 
These models provide scalable geometric priors for in-the-wild inputs and have been increasingly adopted by geometry-aware MLLMs\cite{fan_vlm-3r_2026,wu_spatial-mllm_2025}. Nevertheless, their outputs are primarily designed for reconstruction rather than language-conditioned spatial reasoning, and existing MLLMs typically use them as auxiliary token-level features through projection layers or cross-attention\cite{fan_vlm-3r_2026,wu_spatial-mllm_2025}.
In contrast, our work converts feed-forward geometry into a voxelized cognitive map, where semantic features are aggregated in a persistent 3D metric space to support cross-view and allocentric spatial reasoning.

\subsection{Brain-inspired  spatial reasoning for MLLM}
Brain-inspired reasoning has recently been explored as a way to improve the spatial reasoning of MLLMs. At a high level, cognitive maps offer a useful abstraction for organizing local observations into structured environmental representations~\cite{epstein_cognitive_2017,kuipers_spatial_2000,zhang_theory_2026}.
However, most existing MLLM studies adopt this idea only as an auxiliary reasoning scaffold.
For example, Thinking-in-Space shows that a lightweight grid-based cognitive map can improve MLLMs' spatial layout understanding~\cite{yang_thinking_2024}.
SpaceR encourages 2D object-layout map imagination during reinforcement learning~\cite{ouyang_spacer_2025}, while Video2Layout replaces coarse grid maps with metric-grounded BEV layouts for finer spatial computation~\cite{huang_video2layout_2025}.
These methods highlight the value of spatial layouts, but their representations remain largely planar and object-level abstractions.

Another line of work develops 3D-aware representations for spatial reasoning. 
LLaVA-3D constructs 3D patches by injecting 3D position embeddings into 2D visual patches~\cite{zhu_llava-3d_2025}, while 3DLLM-Mem introduces long-term spatiotemporal memory to retrieve task-relevant information across extended trajectories~\cite{hu_3dllm-mem_2025}.
More recently, Map2Thought constructs an object-centric Metric-CogMap for deterministic Cog-CoT reasoning\cite{gao_map2thought_2026}, while Cog3DMap builds a compact 3D memory from multi-view images and directly feeds map tokens into the MLLM instead of visual tokens\cite{gwak_cog3dmap_2026}. 
In contrast, SpaceMind++ does not treat the map as a final object-level reasoning layout or as a replacement for visual tokens. Instead, it uses the map as an intermediate coordinate-grounded neural workspace and returns spatial knowledge to the original visual-token interface.

\begin{figure*}[t]
\centering
\includegraphics[width=\textwidth]{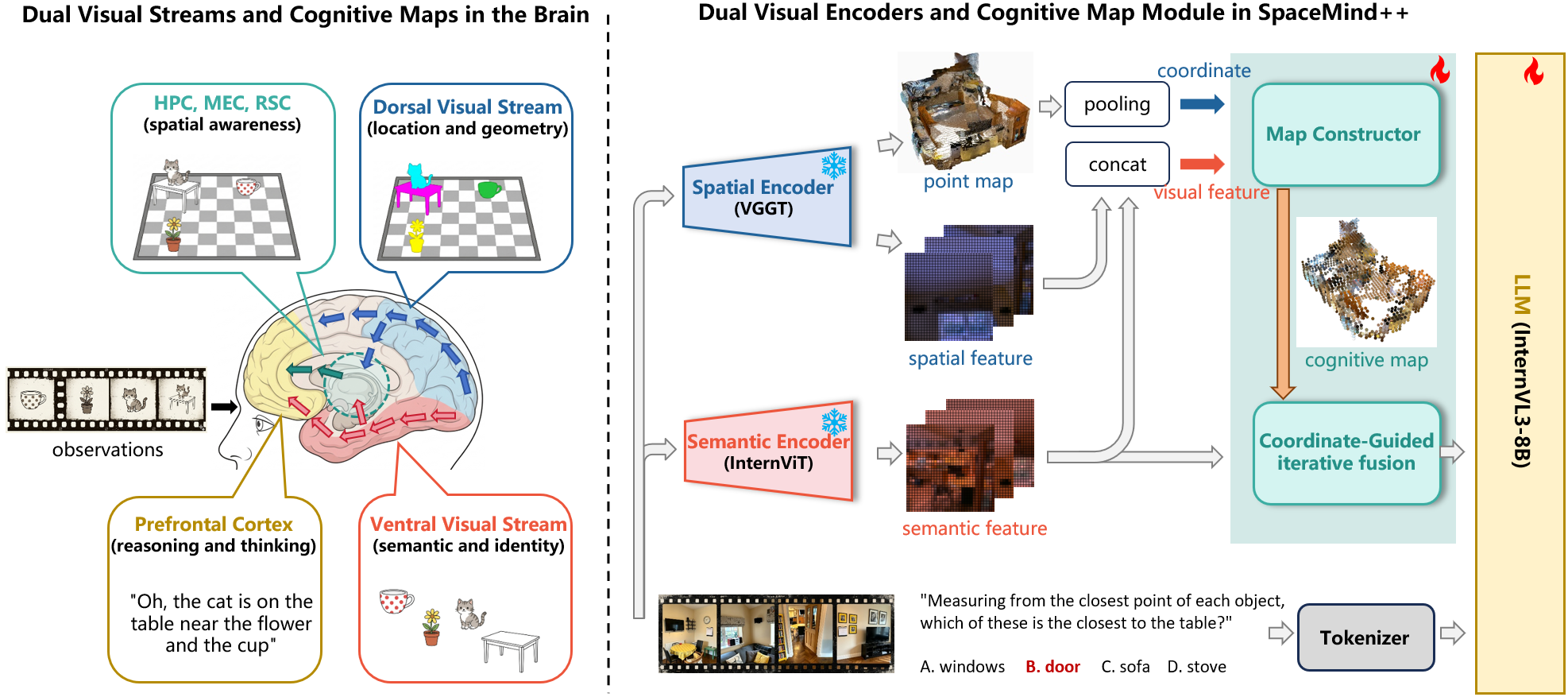}
\caption{
\textbf{Left}: biological motivation. Mammalian spatial cognition separates semantic identity and geometric localization through ventral (red area) and dorsal visual streams (red area), and integrates them into an allocentric cognitive map (cyan area) for spatial awareness and reasoning. 
\textbf{Right}: model architecture. SpaceMind++ extracts semantic and spatial features from video, organizes them into a voxelized allocentric cognitive map, and uses CDIF to relay map-level spatial knowledge back to visual tokens before language decoding (downward yellow arrow).
}
\label{fig:architecture}
\end{figure*}

\section{Method}

\subsection{Overall Architecture}
The goal of SpaceMind++ is to empower Multimodal Large Language Models (MLLMs) with a 3D cognitive structure, enabling precise spatial reasoning in physical environments. Given a sequence of RGB video frames $\mathcal{S} = \{I_i\}_{i=1}^N$, $I_i \in \mathbb{R}^{3 \times H \times W}$ and a textual prompt $T$, our model performs a multi-stage transformation that grounds language reasoning in a structured 3D metric space. SpaceMind++ follows the brain-inspired dual-stream design (left in Figure~\ref{fig:architecture}) and instantiates it as a three-stage video MLLM architecture (right in Figure~\ref{fig:architecture}): dual-stream feature extraction, allocentric cognitive map construction, and map-to-visual fusion.

\paragraph{Dual-stream feature extraction.}
Inspired by the biological dual-pathway process, we first decompose the input video into semantic and geometric streams.

\textbf{Ventral stream} A pretrained 2D visual encoder $E_v$ (InternViT) extracts semantic tokens $f_v \in \mathbb{R}^{N \times M_v \times D_v}$ from the video frames, capturing \textit{what} is present in the scene.

\textbf{Dorsal stream} Simultaneously, a geometry-aware spatial encoder $E_s$ (e.g., VGGT\cite{Wang2025VGGTVG}) processes the same input to produce the geometric latent features $f_s \in \mathbb{R}^{N \times M_v \times D_s}$, per-pixel 3D coordinates $P \in \mathbb{R}^{N \times H \times W \times 3}$ and confidence scores $C \in \mathbb{R}^{N \times H \times W \times 1}$, representing \textit{where} objects are located.
\begin{equation}
f_v = E_v(\mathcal{S}), \quad \{f_s, P, C\} = E_s(\mathcal{S}).
\end{equation}

\paragraph{Allocentric cognitive map fusion.}
The core of SpaceMind++ is the Cognitive Map module, which integrates the dual-stream features in an explicit 3D space. 
It scatters egocentric visual tokens into a unified voxel grid according to their 3D coordinates, producing an allocentric cognitive map $\mathcal{M}$:
\begin{equation}
\mathcal{M} = \mathrm{Voxelize}(f_v, f_s, P),  \quad f_{\mathrm{fused}} = \mathcal{F}(f_v, \mathcal{M}, P).
\end{equation}
To make the map accessible to the pretrained LLM without changing its visual-token interface, $\mathcal{F}$ alternates between \emph{spatial reasoning} over map tokens and \emph{map reading} from map tokens back to visual tokens (the coordinate-guided iterative fusion module in Figure~\ref{fig:architecture}), progressively injecting 3D spatial knowledge into the original visual stream as $f_{\mathrm{fused}}$.

\paragraph{Language decoding.}
The resulting geometry-enhanced visual tokens are fed into the LLM backbone $G$ together with the text prompt $T$ to generate response $R$

\subsection{Allocentric Cognitive Map Construction}

As illustrated in the left part of Figure~\ref{fig:components}, the map constructor converts patch-level visual features and 3D coordinates into a compact voxelized cognitive map through coordinate quantization, topology-preserving aggregation, and stochastic sampling.

\subsubsection{Dynamic Coordinate Quantization}

We first align the dense geometric outputs of the spatial encoder with the patch-level visual tokens. 
Specifically, the dense point map and confidence map are average-pooled to the visual-token resolution, producing patch-aligned coordinates $P \in \mathbb{R}^{L \times 3}$ and confidence scores $C \in \mathbb{R}^{L \times 1}$, where $L=N \times M_v$. 
To reduce the effect of noisy geometry, we retain only high-confidence coordinates and compute a dynamic scene center $\mathbf{c} \in \mathbb{R}^3$ as their centroid:

\begin{equation}
\mathbf{c} = \frac{1}{|\mathcal{V}_{\mathrm{valid}}|} \sum_{i \in \mathcal{V}_{\mathrm{valid}}} P_{i},
\quad \text{where} \quad
\mathcal{V}_{\mathrm{valid}} = \{ i \mid C_{i} > \tau_{\mathrm{conf}} \}.
\end{equation}
Here, $\tau_{\mathrm{conf}}$ is the confidence threshold.
We then recenter each coordinate by the dynamic scene center, $\tilde{P}_{i}=P_i-\mathbf{c}$, and discretize the recentered 3D space into a voxel grid with extent $D$ (we set $D$ as 100 for all experiments) and scale-free resolution $r$.
Since the spatial encoder predicts geometry in a relative metric space, $r$ is defined on this normalized scale rather than in absolute physical units.
The continuous coordinates are quantized into integer voxel indices $\mathbf{u}_{i}=(u_{x,i},u_{y,i},u_{z,i})\in\mathbb{Z}^3$:

\begin{equation}
\mathbf{u}_{i} = \left\lfloor \frac{\tilde{P}_{i}}{r} \right\rfloor + \frac{D}{2}.
\end{equation}
This step corresponds to the feature voxelization panel in Figure~\ref{fig:components}, where continuous 3D coordinates are mapped into discrete voxel indices so that tokens from nearby locations can be grouped together. After that, we convert each 3D voxel index into a unique 1D spatial hash key $h_i$ for efficient grouping:
\begin{equation}
h_i = u_{x,i} D^2 + u_{y,i} D + u_{z,i}.
\end{equation}
Because mapping 3D spatial coordinates into a 1D index array strictly requires a predefined boundary limit, tokens whose computed 3D indices fall outside the grid extent $[0, D-1]^3$ are truncated. 

\subsubsection{Topology-Preserving Feature Aggregation}
In video streams, the same physical region may be observed repeatedly, causing substantial token redundancy.
We therefore use voxelization as a geometry-aware bottleneck and group tokens by their 1D spatial hash keys $h$. Let $\mathcal{H}_j=\{i\mid h_i=j\}$ denote the set of tokens assigned to the $j$-th voxel.

\begin{figure*}[t] 
\centering 
\includegraphics[width=\textwidth]{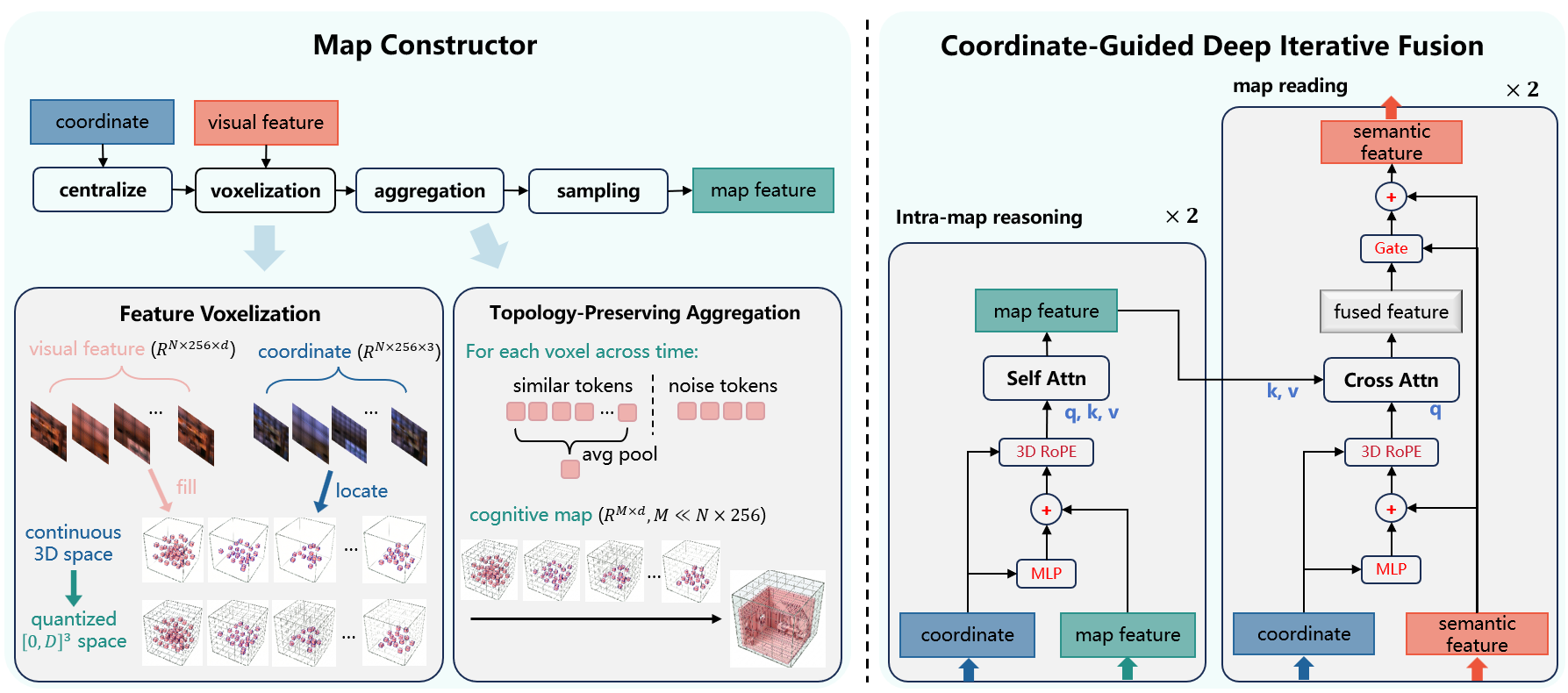} 
\caption{ Detailed components of SpaceMind++. The map constructor transforms patch-level visual features and 3D coordinates into a voxelized cognitive map. CDIF then alternates between intra-map reasoning and map reading to inject spatial knowledge back into semantic tokens. Both stages are guided by coordinate embeddings and 3D RoPE to preserve metric 3D relationships. } 
\label{fig:components} 
\end{figure*}

As shown in Figure~\ref{fig:components}, tokens within each voxel are aggregated into one compact map token after filtering inconsistent observations.
We first compute a voxel centroid $\boldsymbol{\mu}_j$ by averaging the concatenated features $x_i=[f_{v,i};f_{s,i}]$ in the bin, which represents the consensus semantic-geometric state of that physical region.
We then remove outlier observations whose cosine similarity to $\boldsymbol{\mu}_j$ is below $\tau_{\mathrm{sim}}$:

\begin{equation}
\tilde{\mathcal{H}}_j = \left\{ i \in \mathcal{H}_j \,\middle|\, \frac{x_i \cdot \boldsymbol{\mu}_j}{\|x_i\|\,\|\boldsymbol{\mu}_j\|} > \tau_{\mathrm{sim}} \right\}, \quad \boldsymbol{\mu}_j = \frac{1}{|\mathcal{H}_j|} \sum_{i \in \mathcal{H}_j} x_i.
\end{equation}
The final representation for the $j$-th voxel is obtained by applying scatter-mean pooling over the refined set $\tilde{\mathcal{H}}_j$:
\begin{equation}
\mathbf{v}_j = \frac{1}{M_j} \sum_{i \in \tilde{\mathcal{H}}_j} x_i,
\quad
\mathbf{p}_{j} = \frac{1}{M_j} \sum_{i \in \tilde{\mathcal{H}}_j} \tilde{P}_{i},
\quad
t_j = \min_{i \in \tilde{\mathcal{H}}_j} t_i,
\quad
M_j = |\tilde{\mathcal{H}}_j|.
\end{equation}
Here, $\mathbf{v}_j$, $\mathbf{p}_{j}$, and $t_j$ denote the voxel feature, spatial coordinate, and first-observed timestamp of the $j$-th voxel, respectively. 
By retaining the minimum timestamp among all assigned tokens, each voxel records when its corresponding physical region first appears in the video, providing a lightweight cue for appearance-order reasoning. 

\subsubsection{Global Stochastic Sampling for Scale Invariance}
For large scenes, voxelization can yield dense maps.
When the number of voxels $M$ exceeds $M_{\max}$, we uniformly sample $M_{\max}$ valid voxels instead of truncating the sequence, reducing redundancy while preserving the global spatial layout.
We set $M_{\max}=5000$ in all experiments.

\subsection{Coordinate-Guided Deep Iterative Fusion}
After constructing the voxelized cognitive map, SpaceMind++ uses Coordinate-Guided Deep Iterative Fusion (CDIF) to transfer map-level spatial knowledge back to visual tokens while preserving the original visual-token interface. As shown in Figure~\ref{fig:components}, CDIF consists of $L$ layers that alternate between Map Reasoning and Map Reading.

\subsubsection{Map Reasoning via Self-Attention}
Within each fusion layer $l$, CDIF first performs intra-map self-attention to propagate semantic and geometric information across voxels.
We encode each voxel coordinate $\mathbf{p}_{j}$ with an MLP and add it to the voxel feature $\mathbf{v}_{j}^{(l)}$, then apply 3D Continuous Rotary Positional Embedding (3D RoPE)~\cite{su_roformer_2021} to make the queries and keys coordinate-aware.
This enables attention to model relative spatial relationships in metric 3D space:
\begin{equation}
\begin{aligned}
\tilde{\mathbf{v}}_{j}^{(l)} &= \mathbf{v}_{j}^{(l)} + \mathrm{MLP}(\mathbf{p}_{j}), \quad
\hat{\mathbf{q}}_{j}^{(l)}, \hat{\mathbf{k}}_{j}^{(l)} &= \mathrm{RoPE}(\mathrm{Linear}(\tilde{\mathbf{v}}_{j}^{(l)}), \mathbf{p}_{j}).
\end{aligned}
\end{equation}
The voxel map is then updated via self-attention:
\begin{equation}
\mathcal{V}^{(l+1)} = \mathrm{SelfAttn}(\hat{\mathbf{Q}}, \hat{\mathbf{K}}, \mathbf{V}(\mathcal{V}^{(l)})).
\end{equation}
This step enables the cognitive map to perform global geometric inference, such as modeling object-to-object spatial relationships, within a unified 3D coordinate frame.

\subsubsection{Map Reading via Semantic-Aware Gated Attention}

After the cognitive map is refined, each 2D visual token reads spatial context from the map. 
For a visual token $\mathbf{f}_{v,i}^{(l)}$ with 3D coordinate $\mathbf{p}_{i}$, we first inject its coordinate information through an MLP-based embedding. 
The coordinate-enhanced token is then projected into query and key modulated by 3D RoPE, making the map-reading attention aware of its metric location:

\begin{equation}
\tilde{\mathbf{f}}_{v,i}^{(l)} = \mathbf{f}_{v,i}^{(l)} + \mathrm{MLP}(\mathbf{p}_{i}), \quad \mathbf{q}_{i}^{(l)} = \mathrm{RoPE}(\mathrm{Linear}(\tilde{\mathbf{f}}_{v,i}^{(l)}), \mathbf{p}_{i}).
\end{equation}
The model then performs cross-attention over the refined voxel map $\mathcal{V}^{(l+1)}$ to retrieve the fused spatial feature $\mathbf{f}_{\mathrm{fused},i}^{(l)}$:
\begin{equation}
\mathbf{f}_{\mathrm{fused},i}^{(l)} = \mathrm{CrossAttn}(\mathbf{q}_{i}^{(l)}, \mathbf{K}(\mathcal{V}^{(l+1)}), \mathbf{V}(\mathcal{V}^{(l+1)})).
\end{equation}

\paragraph{Gated information integration.}
To prevent the visual stream from absorbing irrelevant geometric background, we introduce a visual-driven gating mechanism. Instead of unconditionally integrating all spatial information, the model uses the semantic content of the visual tokens to control the reading intensity. The gate $\mathbf{g}_{i}^{(l)}$ is computed solely from the visual feature:
\begin{equation}
\mathbf{g}_{i}^{(l)} = \sigma(\mathrm{MLP}(\mathbf{f}_{v,i}^{(l)})).
\end{equation}
The final updated visual representation is then formulated as:
\begin{equation}
\mathbf{f}_{v,i}^{(l+1)} = \mathbf{f}_{v,i}^{(l)} + \mathbf{g}_{i}^{(l)} \odot \mathrm{FFN}(\mathbf{f}_{\mathrm{fused},i}^{(l)}).
\end{equation}
This design ensures that the 2D visual stream remains the primary carrier of semantic identity while selectively integrating 3D spatial knowledge only when the visual content, such as a specific object or landmark, requires geometric grounding for downstream reasoning.

\section{Implementation Details}

\begin{wrapfigure}{r}{0.42\textwidth}
\vspace{-4pt}
\centering
\includegraphics[width=0.41\textwidth]{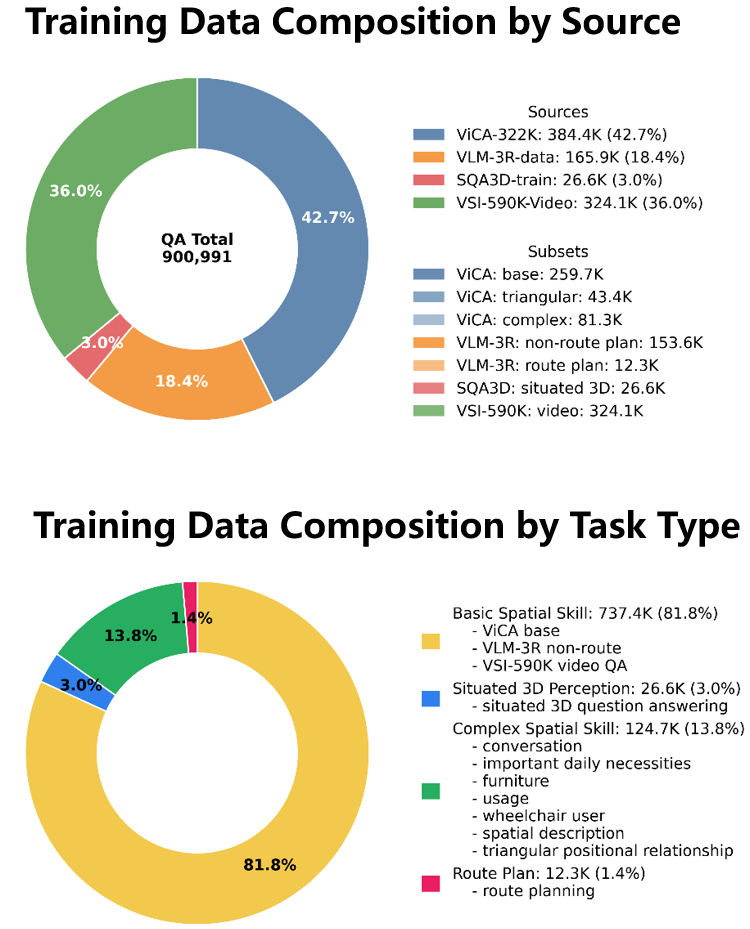}
\caption{
Statistics of the SpaceMind-900K datasets
}
\label{fig:data-mixture}
\vspace{-4pt}
\end{wrapfigure}

\paragraph{Training data.}
We train SpaceMind++ on a spatial instruction-tuning corpus containing approximately 900k QA samples. The corpus integrates four 3D reasoning sources, including ViCA-322K\cite{feng_towards_2025}, VLM-3R-data\cite{fan_vlm-3r_2026}, SQA3D-train\cite{ma_sqa3d_2023}, and VSI-590K-Video\cite{yang_cambrian-s_2025}. As shown in Figure~\ref{fig:data-mixture}, the dataset covers a broad spectrum of spatial reasoning skills, including object counting, relative and absolute distance estimation, size estimation, object ordering, situated 3D perception, complex spatial reasoning, and route planning. This mixture encourages the model to learn both local metric grounding and global spatial consistency, which are essential for reasoning on an allocentric cognitive map.

\paragraph{Training setup.}
We use InternVL3-8B~\cite{zhu_internvl3_2025} as the base MLLM and keep both the ViT backbone and VGGT encoder frozen during training.
We train the CDIF layers and the associated cross-modal projectors, while applying LoRA~\cite{Hu2021LoRALA} to the language model with rank $r=256$ and scaling factor $\alpha=512$.
The trainable parameters are optimized with AdamW~\cite{Loshchilov2017DecoupledWD} using a cosine annealing schedule and a peak learning rate of $2\times10^{-5}$.
Training is conducted on 64 NVIDIA H200 GPUs for approximately 41 hours.

\section{Evaluation}
\begin{table*}[t]
\centering
\scriptsize
\setlength{\tabcolsep}{3.5pt}
\caption{
VSI-Bench evaluation. 
SpaceMind++ achieves the best overall performance and obtains the highest scores on most sub-tasks, demonstrating strong and balanced spatial reasoning ability.
}
\label{tab:vsibench}
\resizebox{\textwidth}{!}{%
\begin{tabular}{l|c|cccc|cccc}
\toprule
\multirow{2}{*}{Methods} & \multirow{2}{*}{Avg.} & \multicolumn{4}{c|}{Numerical Question} & \multicolumn{4}{c}{Multiple-Choice Question} \\
\cmidrule(lr){3-6} \cmidrule(lr){7-10}
 & & Obj. Cnt. & Abs. Dist. & Obj. Size & Room Size & Rel. Dist. & Rel. Dir. & Route Plan & Appr. Order \\
\midrule

\multicolumn{10}{l}{\textit{Proprietary Models (API)}} \\
GPT-5~\cite{openai_gpt5_2026} & \underline{55.0} & \textbf{53.3} & 34.4 & \textbf{73.3} & \underline{47.5} & \textbf{63.7} & \underline{48.6} & \underline{50.2} & \textbf{68.9} \\
Gemini-3 Pro~\cite{google_gemini3_pro_2025} & \textbf{56.0} & \underline{49.0} & \textbf{42.8} & \underline{71.5} & 41.8 & 56.6 & \textbf{57.5} & \textbf{61.9} & 60.0 \\
Grok-4~\cite{xai_grok4_2025} & 47.9 & 37.1 & 32.9 & 60.8 & 45.4 & 53.1 & 39.6 & 47.4 & 66.8 \\

\midrule
\multicolumn{10}{l}{\textit{Open-source VLMs}} \\
InternVL3-78B~\cite{zhu_internvl3_2025} 
& \underline{48.5} & \textbf{71.2} & \textbf{53.7} & 44.4 & \underline{39.5} & \underline{55.9} & 39.5 & 28.9 & \underline{54.5} \\
LLaVA-NeXT-Video-72B~\cite{liu_llava-next_2024} 
& 40.9 & 48.9 & 22.8 & 57.4 & 35.3 & 42.4 & 36.7 & \textbf{35.0} & 48.6 \\
Qwen3VL-8B-Instruct~\cite{yang_qwen3_2025} 
& \textbf{57.9} & \underline{67.5} & \underline{47.0} & \textbf{76.3} & \textbf{61.9} & \textbf{58.0} & \textbf{50.9} & \textbf{35.0} & \textbf{66.3} \\
LLaVA-OneVision-72B~\cite{li_llava-onevision_2024} 
& 40.2 & 43.5 & 23.9 & \underline{57.6} & 37.5 & 42.5 & \underline{39.9} & 32.5 & 44.6 \\

\midrule
\multicolumn{10}{l}{\textit{Specialized Spatial Reasoning Models}} \\
Spatial-MLLM~\cite{wu_spatial-mllm_2025} 
& 48.4 & 65.3 & 34.8 & 63.1 & 45.1 & 41.3 & 46.2 & 33.5 & 46.3 \\
VST-7B~\cite{yang_visual_spatial_tuning_2025}
& 60.6 & 72.0 & 44.4 & 74.3 & 68.3 & 59.7 & 55.8 & 44.9 & 65.2 \\
VLM-3R~\cite{fan_vlm-3r_2026} 
& 60.9 & 70.2 & 49.4 & 69.2 & 67.1 & 65.4 & 80.5 & 45.4 & 40.1 \\
SenseNova-SI-1.2~\cite{cai_scaling_spatial_intelligence_2025} 
& 69.6 & 72.7 & 56.0 & 77.1 & \underline{75.7} & \underline{70.4} & 81.7 & 42.8 & \underline{79.9} \\
SpaceMind~\cite{zhao_spacemind_2025} 
& \underline{70.2} & \underline{73.9} & \underline{61.5} & \underline{77.6} & 74.8 & 67.7 & \underline{88.6} & \underline{46.9} & 70.7 \\
\rowcolor{gray!12}
\textbf{SpaceMind++ (Ours)} 
& \textbf{73.2} & \textbf{74.5} & \textbf{62.4} & \textbf{77.9} & \textbf{76.9} & \textbf{73.5} & \textbf{89.7} & \textbf{48.2} & \textbf{84.1} \\
\bottomrule
\end{tabular}%
}
\end{table*}
We evaluate SpaceMind++ on four spatial reasoning benchmarks.
VSI-Bench is used as the primary benchmark for video-based spatial intelligence, while SPBench, SITE-Bench, and SPAR-Bench are used to assess generalization across multi-view and video spatial reasoning settings.

\subsection{VSI-Bench}
As shown in Table~\ref{tab:vsibench} and Figure~\ref{fig:data-mixture}, SpaceMind++ achieves a new state-of-the-art average score of 73.2, improving upon the strong SpaceMind baseline on VSI-Bench. 
Although SpaceMind and SenseNova-SI are already highly competitive spatial reasoning models, SpaceMind++ further improves several key subcategories that directly reflect the benefits of allocentric cognitive map construction.

Specifically, SpaceMind++ improves relative distance estimation from 67.7 to 73.5 and appearance
ordering from 70.7 to 84.1. Relative distance estimation requires metric grounding between objects in 3D space, while appearance ordering further requires preserving temporal object events and spatial relationships across changing viewpoints. We also observe competitive performance on object count, room size, and relative direction estimation. 
Overall, the performance suggests that SpaceMind++ is most effective when spatial reasoning requires persistent object-level memory and metric consistency across viewpoints, which are precisely the capabilities targeted by the voxelized cognitive map.

\subsection{Generalization to Spatial Reasoning Benchmarks}

\paragraph{SPBench.}
\begin{table*}[t]
\centering
\scriptsize
\setlength{\tabcolsep}{4.6pt}

\caption{
Generalization evaluation on SPBench and SITE-Bench.
SpaceMind++ achieves strong out-of-domain performance, especially on multi-view SPBench and video-only SITE-Bench, demonstrating its advantage in cross-view and video spatial reasoning.
}

\label{tab:generalization}
\begin{tabular}{l|ccc|ccc|c|c}
\toprule
\multirow{2}{*}{Methods} 
& \multicolumn{3}{c|}{SPBench-SI} 
& \multicolumn{3}{c|}{SPBench-MV} 
& \multirow{2}{*}{SPBench}
& \multirow{2}{*}{SITE-Bench} \\
\cmidrule(lr){2-4} \cmidrule(lr){5-7}
& NQ & MCQ & Avg. & NQ & MCQ & Avg. & & \\
\midrule
\multicolumn{9}{l}{\textit{Proprietary Models}} \\
GPT-4o~\cite{openai_gpt-4o_2024}
& 24.5 & 60.3 & 42.4 & 40.7 & 59.4 & 50.1 & 46.2 & -- \\
Gemini-2.0-Flash~\cite{comanici_gemini_2025}
& 49.0 & 60.4 & 54.7 & 51.9 & 50.7 & 51.3 & 53.0 & -- \\

\midrule
\multicolumn{9}{l}{\textit{Open-Source Models}} \\
InternVL-2.5-8B~\cite{chen_internvl25_2024}
& 28.3 & 56.3 & 42.3 & 37.3 & 47.5 & 42.4 & 42.3 & -- \\
Kimi-VL-A3B~\cite{moonshotai_kimivl_2025}
& 25.7 & 44.9 & 35.3 & 23.3 & 57.6 & 40.5 & 37.9 & -- \\
LLaVA-OneVision-7B~\cite{li_llava-onevision_2024}
& 25.4 & 41.0 & 33.2 & 20.6 & 49.6 & 35.1 & 34.2 & -- \\
Qwen2.5-VL-7B~\cite{bai_qwen25-vl_2025}
& 36.3 & 60.5 & 48.4 & 28.9 & 49.8 & 39.3 & 43.9 & 53.7 \\

\midrule
\multicolumn{9}{l}{\textit{Spatial Reasoning Models}} \\
Video-R1~\cite{feng_video-r1_2025}
& 27.7 & 62.0 & 44.9 & 32.5 & 53.0 & 42.8 & 43.8 & -- \\
SpaceR-7B~\cite{ouyang_spacer_2025}
& 35.7 & 61.5 & 48.6 & 63.2 & 53.7 & 58.5 & 53.5 & 56.5 \\
VILASR-7B~\cite{wu_reinforcing_2025}
& 36.6 & \underline{63.7} & 50.2 & 56.2 & 59.6 & 57.9 & 54.0 & 56.1 \\
Spatial-MLLM-4B~\cite{wu_spatial-mllm_2025}
& 38.1 & 49.3 & 43.7 & 63.7 & 58.9 & 61.3 & 52.5 & 44.0 \\
SpaceMind~\cite{zhao_spacemind_2025}
& \textbf{66.3} & 53.2 & \underline{59.7} & \underline{76.2} & \underline{70.5} & \underline{73.8} & \underline{67.3} & -- \\
EgoMind~\cite{chen_egomind_2026}
& -- & -- & -- & -- & -- & -- & 55.0 & \underline{58.0} \\
\rowcolor{gray!12}
\textbf{SpaceMind++ (Ours)}
& \underline{56.7} & \textbf{65.6} & \textbf{61.1}
& \textbf{76.5} & \textbf{82.3} & \textbf{78.9}
& \textbf{70.0} & \textbf{61.0} \\
\bottomrule
\end{tabular}
\end{table*}
SPBench evaluates spatial reasoning under both single-image and multi-view (eight
images) settings\cite{li_spatialladder_2025}. As shown in Table~\ref{tab:generalization}, SpaceMind++ maintains competitive performance on the single-image subset while achieving a substantial improvement on the multi-view subset. Specifically, SpaceMind++ obtains 61.1 on SPBench-SI and 78.9 on SPBench-MV, leading to a strong overall score. The large gain on SPBench-MV is particularly consistent with our design motivation: by projecting
frame-level observations into a shared 3D cognitive map, SpaceMind++ can better aggregate cross-view evidence and maintain object relationships across viewpoints. This suggests that the learned
spatial representation generalizes to out-of-domain multi-view spatial QA.
\paragraph{SITE-Bench.}
SITE-Bench evaluates spatial intelligence in a multiple-choice VQA format across single-image, multi-image, and video inputs\cite{wang_site_2025}. Since its image subset mostly contains only 1--5 input frames, such short visual contexts provide limited cross-view evidence for constructing a reliable cognitive map.
We only report the video-only result under the EgoMind-style average metric as an auxiliary generalization reference and do not claim full SITE-Bench superiority. As shown in Table~\ref{tab:generalization}, SpaceMind++ achieves competitive performance on SITE-Bench.

\begin{table*}[t]
\centering
\scriptsize
\setlength{\tabcolsep}{4.9pt}
\caption{
Generalization evaluation on SPAR-Bench high-level tasks.
SpaceMind++ achieves the best average performance, with strong results on multi-view (MV) object-relation (ObjRel) and spatial-imagination (SpImag) tasks, highlighting its advantage in cross-view spatial consistency.
}
\label{tab:sparbench_high}
\begin{tabular}{l|c|ccccccccc}
\toprule
Methods 
& \makecell{High\\Avg.}
& \makecell{Dist\\OO}
& \makecell{Dist\\OO-MV}
& \makecell{ObjRel\\OC-MV}
& \makecell{ObjRel\\OO}
& \makecell{ObjRel\\OO-MV}
& \makecell{SpImag\\OC}
& \makecell{SpImag\\OC-MV}
& \makecell{SpImag\\OO}
& \makecell{SpImag\\OO-MV} \\
\midrule
\multicolumn{11}{l}{\textit{Closed-source Models}} \\
GPT-4o~\cite{openai_gpt-4o_2024} 
& 43.80 & 65.00 & 64.88 & 44.75 & 50.82 & 43.21 & 29.84 & 32.56 & 27.81 & 35.29 \\
GPT-4.1~\cite{openai_gpt4_2023} 
& 42.93 & \underline{71.76} & \underline{67.26} & 46.25 & 54.95 & 41.00 & 30.38 & 29.65 & 20.20 & 24.93 \\
Doubao-1.5-vision-pro~\cite{seed_team_seed15vl_2025} 
& \underline{49.49} & \textbf{74.71} & \textbf{69.64} & 35.75 & \textbf{70.33} & \underline{47.65} & 34.95 & 33.14 & \textbf{35.76} & \underline{42.86} \\
\midrule
\multicolumn{11}{l}{\textit{Open-source Models}} \\
InternVL2.5-38B~\cite{chen_internvl25_2024} 
& 44.13 & 69.12 & 66.67 & 43.75 & \underline{64.29} & 37.67 & 25.27 & 31.98 & 31.79 & 26.61 \\
Qwen2.5-VL-72B~\cite{bai_qwen25-vl_2025} 
& 43.80 & 58.82 & 61.90 & 40.75 & 53.57 & 45.98 & 26.88 & 35.17 & \underline{34.11} & 36.97 \\
LLaVA-v1.6-7B~\cite{liu_llava-next_2024} 
& 20.18 & 51.76 & 7.74 & 6.25 & 32.14 & 6.37 & \underline{39.52} & 10.47 & 21.52 & 5.88 \\
SpaceR-7B~\cite{ouyang_spacer_2025} 
& 45.61 & 62.35 & 61.61 & \underline{52.25} & 51.92 & 46.81 & 37.90 & \underline{36.05} & 24.83 & 34.17 \\
\rowcolor{gray!12}
\textbf{SpaceMind++ (Ours)} 
& \textbf{51.28} & 39.78 & 33.00 & \textbf{70.00} & 48.63 & \textbf{53.46} & \textbf{51.61} & \textbf{53.78} & 27.15 & \textbf{45.66} \\
\bottomrule
\end{tabular}
\end{table*}
\paragraph{SPAR-Bench.}
SPAR-Bench provides a difficulty-stratified evaluation of spatial understanding\cite{zhang_flatland_2025}. 
Here, we focus on the high-level split, which emphasizes multi-object relational reasoning and spatial imagination across views. 
We omit the low- and medium-level splits from the main table because they are dominated by direct depth, distance, and view-alignment perception, while our method is primarily designed to improve persistent spatial representation and cross-view relational reasoning. 
As shown in Table~\ref{tab:sparbench_high}, SpaceMind++ achieves the best average performance among open-source models, with clear advantages on multi-view object-relation and spatial-imagination tasks.
These tasks require cross-view object association and hypothetical-view reasoning, where our voxelized cognitive map provides a stable 3D reference frame for object identities and spatial relations.

\subsection{Ablation Studies}
We conduct ablation studies on VSI-Bench to examine the effects of the spatial encoder, cognitive-map construction, and coordinate-guided iterative fusion.
As shown in Table~\ref{tab:ablation}, the spatial-encoder baseline and architecture-ablation groups are trained on an earlier 626K corpus, which does not include the VSI-590K data, while the final SpaceMind++ model is trained on the final 900K corpus.
The architecture ablations use an InternVL3-8B and InternVL3.5-8B backbone. Therefore, this table is intended to reveal component-level trends rather than provide a strictly controlled comparison under identical training settings.
\begin{wraptable}{r}{0.48\textwidth}
\centering
\vspace{-2pt}
\scriptsize
\setlength{\intextsep}{0.5pt}
\setlength{\tabcolsep}{3.5pt}
\caption{
Ablation study on VSI-Bench. 
}
\label{tab:ablation}
\begin{tabular}{l|ccc}
\toprule
Methods & Avg. & Num. & MCQ \\
\midrule
\multicolumn{4}{l}{\textit{Spatial-Encoder Baseline}} \\
InternVL3-8B SFT (626K)& 63.7 & 65.5 & 61.9 \\
InternVL3-8B + VGGT (626K)& 64.6 & 66.2 & 63.0 \\
\midrule
\multicolumn{4}{l}{\textit{Model Architecture}} \\
InternVL3.5-8B + DIF (626K)& 66.3 & 67.5 & 65.0 \\
InternVL3.5-8B + CogMap + DIF (626K)& 68.9 & 69.5 & 68.2 \\
\rowcolor{gray!12}
\textbf{SpaceMind++} (900K)& \textbf{73.2} & \textbf{72.9} & \textbf{73.9} \\
\bottomrule
\end{tabular}
\vspace{-2pt}
\end{wraptable}

Adding VGGT-based spatial features improves the InternVL3-8B SFT baseline from 63.7 to 64.6, indicating that feed-forward geometry provides useful but limited spatial cues. DIF denotes the non-coordinate-guided variant of our CDIF module.
The architecture ablation further shows that introducing a cognitive map into DIF improves the average score from 66.3 to 68.9, suggesting that organizing visual observations into a structured map is beneficial for spatial reasoning.
The full SpaceMind++ model achieves the best overall performance, reaching 73.2 with CDIF.
This gain should be attributed to both the proposed coordinate-guided architecture, where coordinate embeddings and 3D RoPE explicitly guide information transfer during map reasoning and map-to-visual reading, and the richer 900K training corpus, which includes VSI-590K.

\section{Conclusion}

SpaceMind++ achieves state-of-the-art performance on VSI-Bench and generalizes well to SPBench, SITE-Bench, and SPAR-Bench, with clear gains on tasks requiring object-level memory, metric consistency, and cross-view integration.
These improvements come from treating 3D geometry not as an auxiliary feature, but as an organizing structure: SpaceMind++ consolidates fragmented egocentric observations into a persistent voxelized allocentric cognitive map and transfers spatial knowledge back to visual tokens through Coordinate-Guided Deep Iterative Fusion.
Overall, our results highlight cognitive map construction as a promising direction for spatially grounded video MLLMs.

\paragraph{Limitations and Future Work.}
SpaceMind++ has several limitations.
First, its cognitive map depends on the quality of the feed-forward spatial encoder; inaccurate point maps, incomplete geometry, or sparse observations can weaken the resulting spatial representation.
Second, although voxelization reduces token redundancy, map reasoning and map-to-visual fusion introduce extra computation, and a fixed voxel budget may discard fine-grained details in large scenes.
Future work could explore uncertainty-aware map construction, adaptive or hierarchical voxel allocation, and extensions to dynamic scenes, interactive embodied agents, and long-horizon spatial memory.

\clearpage
\bibliographystyle{plainnat}
\bibliography{uploads/paper_reference}

\end{document}